\documentclass{article}

\usepackage{arxiv}

\usepackage{xspace}
\usepackage{cite}
\usepackage{soul}
\usepackage{amsmath,amssymb,amsfonts}
\usepackage{algorithmic}
\usepackage{graphicx}
\usepackage{textcomp}
\usepackage{graphicx}
\usepackage[table,xcdraw]{xcolor}
\usepackage{multirow}
\usepackage{ulem}
\usepackage{booktabs}
\usepackage{siunitx}
\usepackage{placeins}
\usepackage{subcaption}

\def\BibTeX{{\rm B\kern-.05em{\sc i\kern-.025em b}\kern-.08em
    T\kern-.1667em\lower.7ex\hbox{E}\kern-.125emX}}
\def\mobilenet/{MobileNetV2}
\def\paste/{PaSTe}
\def\resnet/{WideResNet50}

\begin{document}
\title{PaSTe: Improving the Efficiency of \\ Visual Anomaly Detection at the Edge}

\author{
    Manuel Barusco \\ 
    University of Padova, Italy \\ 
    \texttt{manuel.barusco@phd.unipd.it} \\ \And 
    Francesco Borsatti \\ 
    University of Padova, Italy \\ 
    \texttt{francesco.borsatti.1@phd.unipd.it} \\ \And
    Davide Dalle Pezze \\ University of Padova, Italy \\ 
    \texttt{davide.dallepezze@unipd.it} \\ \And
    Francesco Paissan \\ Fondazione Bruno Kessler, Italy \\ 
    \texttt{fpaissan@fbk.eu} \\ \And
    Elisabetta Farella \\ Fondazione Bruno Kessler, Italy \\ 
    \texttt{efarella@fbk.eu} \\ \And
    Gian Antonio Susto \\ 
    University of Padova, Italy \\ 
    \texttt{gianantonio.susto@unipd.it} \\
}

\maketitle

\begin{abstract}
Visual Anomaly Detection (VAD) has gained significant research attention for its ability to identify anomalous images and pinpoint the specific areas responsible for the anomaly. A key advantage of VAD is its unsupervised nature, which eliminates the need for costly and time-consuming labeled data collection. However, despite its potential for real-world applications, the literature has given limited focus to resource-efficient VAD, particularly for deployment on edge devices.
This work addresses this gap by leveraging lightweight neural networks to reduce memory and computation requirements, enabling VAD deployment on resource-constrained edge devices. We benchmark the major VAD algorithms within this framework and demonstrate the feasibility of edge-based VAD using the well-known MVTec dataset. Furthermore, we introduce a novel algorithm, Partially Shared Teacher-student (\paste/), designed to address the high resource demands of the existing Student Teacher Feature Pyramid Matching (STFPM) approach. Our results show that \paste/ decreases the inference time by 25\%, while reducing the training time by 33\% and peak RAM usage during training by 76\%. These improvements make the VAD process significantly more efficient, laying a solid foundation for real-world deployment on edge devices.
\end{abstract}


\section{Introduction}
Visual Anomaly Detection (VAD) is a computer vision task that aims to identify images containing anomalies and pinpoint the specific pixels within the image responsible for the anomaly (see examples in Fig. \ref{Fig:MVTec_Dataset_AD}). This is performed using the unsupervised learning paradigm, avoiding the costly label collection phase necessary for pixel-level anomaly tagging.

VAD has many applications in various fields, such as manufacturing, medicine, and autonomous vehicles \cite{MVTec, bao2023bmad, chan2021segmentmeifyoucan}. However, its relevance is limited by the constraint of its deployment in real-world environments. Most of the current literature focuses on VAD performance as the only important metric, excluding other practical considerations regarding memory, inference time, and processing power. However, in most real-world application scenarios, it is not unusual that VAD algorithms run on edge devices with limited resources, making it challenging to deploy complex deep learning models typically used in VAD, such as \resnet/ \cite{zagoruyko2016wide}.

In this work, we provide a benchmark for resource-efficient VAD (also known as tinyAD) by testing the most well-known VAD methods in the literature, considering relevant metrics for edge deployment.
This is crucial for real-time applications and deployments in environments where resources are limited.
To perform this study, we consider lightweight networks that enable the implementation of VAD on edge devices (see Fig. \ref{fig:comparison_generic}), these networks are often constrained in terms of processing power, memory, and energy consumption. 

Moreover, we address the method's weaknesses by proposing a new algorithm called Partially Shared Teacher-student (\paste/). 
This new algorithm is based on the Student-Teacher Feature Pyramid (STFPM) \cite{st_pyramid} approach and is intended for edge deployment. It aims to run on tiny devices by reducing the required computational resources.
With our approach, applied to four different resource-efficient backbones, we achieve up to 87\% reduction in training memory and 50\% reduction in computation compared to the original STFPM. The trade-off is a slight decrease in performance on some backbones, while others remain unaffected or even improve.
We prove the feasibility of deploying AD methods on the edge and the superiority of \paste/ over STFPM by evaluating with the MVTEC dataset, the most well-known VAD dataset, which consists of ten objects and five textures.

Our contributions can be summarized as follows:
\begin{itemize}
    \item We test several edge architectures in the context of VAD;
    \item We propose a novel AD algorithm specifically designed for the edge, called \paste/; 
    \item We compare several edge architectures and VAD methods, providing a benchmark for resource-efficient VAD by evaluating using the well-known MVTec dataset.
\end{itemize}

The outline of the paper is as follows. 
In Section \ref{sec:related_work}, we describe the VAD algorithms present in the literature and the relevant developments in edge-oriented neural networks.
In Section \ref{sec:methodology}, we introduce the proposed framework for tinyAD and the specific method, \paste/, proposed to reduce resource consumption on the edge of STFPM.
In Section \ref{sec:experimental_setting}, we describe the experimental setup for all the AD methods and tiny backbones considered.
Finally, in Section \ref{sec:results}, we present our findings before concluding in Section \ref{sec:conclusion}.

\begin{figure}[thbp]
  \centering
    \includegraphics[width=0.6\linewidth]{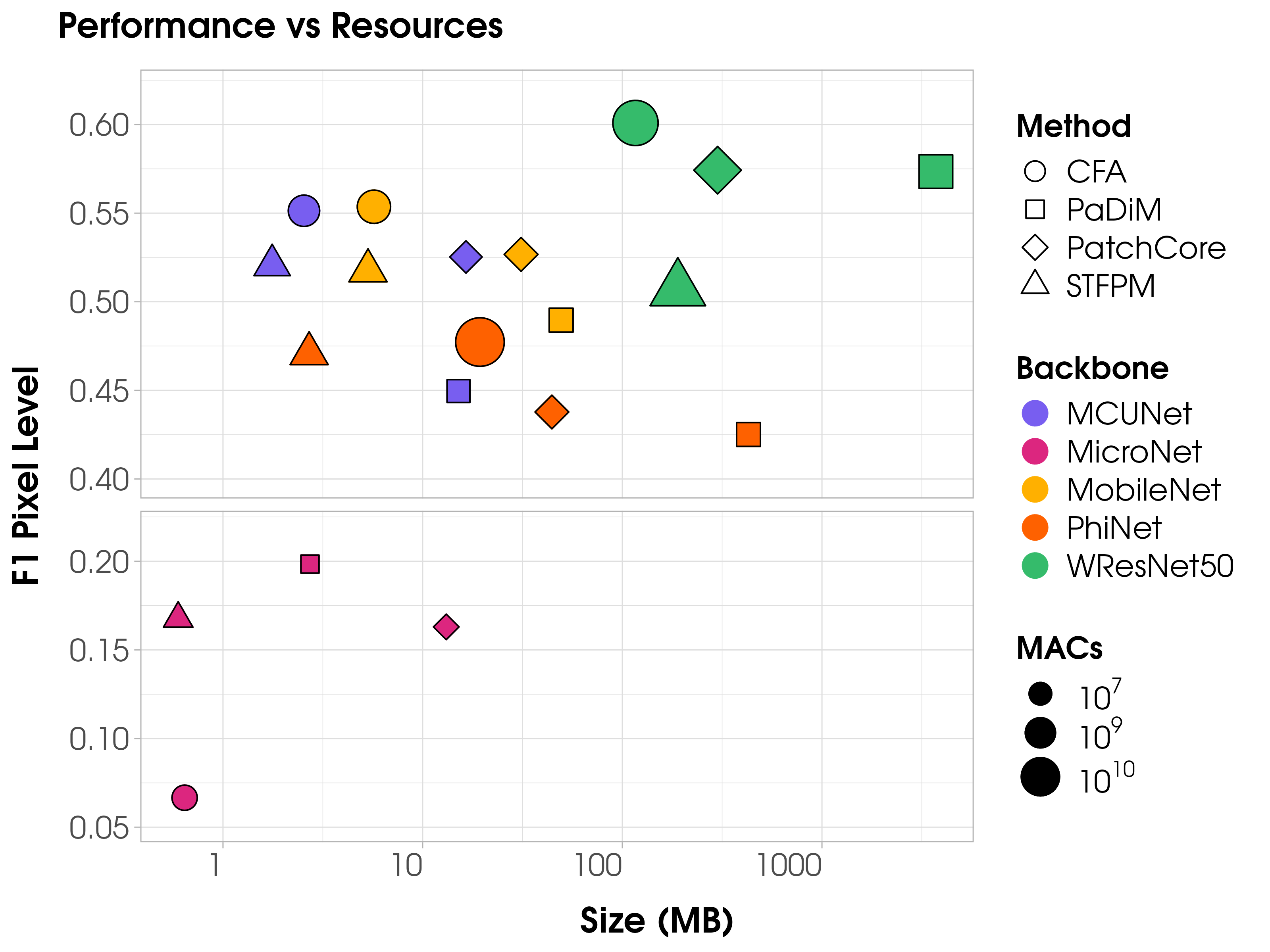}
    \caption{We show on the x-axis the inference time and on the y-axis, the performance. Each color represents a different AD method, while each symbol represents a different tiny backbone. The size represents the total memory required.  }
    \label{fig:performance_tinyAD}
\end{figure}

\section{Related Work}
\label{sec:related_work}

\subsection{Visual Anomaly Detection}
\label{sec:related_work_ad}
AD approaches find many applications in Computer Vision (CV) where safety is crucial, encompassing manufacturing, the medical domain, autonomous vehicles, security systems, and more \cite{MVTec, bao2023bmad, chan2021segmentmeifyoucan}.
In fact, identifying anomalous samples helps users in their decision-making process.
In addition, recent approaches focused on providing interpretability.
This is achieved by enhancing the model predictions from image-level to fine-grained, pixel-level detail. 
Ensuring the interpretability of these systems can lead to safer and more efficient operations in various fields.
Moreover, those approaches focus on the unsupervised paradigm, eliminating the need for a laborious label collection phase.
This phase is typically time-consuming and resource-intensive, requiring substantial human effort and expertise.
We can split most of the approaches into two main families: reconstruction-based methods and Feature-based methods \cite{bao2023bmad, xie2024iad}.
\textbf{Reconstruction-based methods} learn to reconstruct normal images during training using generative models such as AutoEncoders and GANs  \cite{MVTec,akcay2019ganomaly,ZAVRTANIK2021107706,draem}.
These methods rely on identifying anomalies based on the difference between the original and reconstructed images.
However, working in the image domain instead of the feature domain can be extremely expensive in terms of processing power and architecture size (since we move from classification to generative models). Moreover, their performance is usually lower than feature-based methods \cite{xie2024iad}.

In contrast, many proposed state-of-the-art approaches belong to the \textbf{feature-based} family.
Instead of working on the image domain, they consider embedding representations of images produced by a pre-trained model.
Then the feature map produced is divided into patches. Analyzing each region separately helps identify local anomalies. 
 
These approaches can be further categorized as (i) Teacher-Student based, (ii) Normalizing Flow, and (iii) Memory Bank.
\\
\textbf{Memory Bank approaches} capture the features of normal images and store them in a memory bank \cite{PaDiM,patch,lee2022cfa}. Belonging to the category, three approaches are studied: Padim, PatchCore, and CFA.
While these methods show remarkable performance, they require an additional memory, which can be extremely large.
\textbf{PatchCore} filters the normal patches and saves this small portion in memory.
During inference, each patch's test image checks its similarity to the memory patches. The farther the distance, the higher the anomaly score.
In \textbf{Padim}, each patch position of the image is represented by a multivariate gaussian distribution.
During the inference process for a given test image, the Mahalanobis distance is computed for each patch, providing the anomaly score.
\textbf{CFA} constructs a memory of patches and then seeks to enhance the concentration of normal features around these stored patch representations. 
This should help to increase the distance between normal and abnormal patches.

\textbf{Teacher-Student approaches}, as the name suggests, are based on two networks, a teacher and a student architecture.
For example, the STFPM approach takes advantage of the knowledge distillation approach to transfer learned knowledge from teacher to student, and when the features deviate, it is assumed that there is an anomaly \cite{st_pyramid} (see Fig. \ref{fig:STFPM_scheme}).
A disadvantage of these methods is that they require additional memory to store a student network.
\\
\textbf{One-class classification methods} try to learn a representation of normal data during training, usually applying self-supervised techniques.
For example, PatchSVDD applies an encoder to aggregate similar normal patches. However, it requires memorizing all normal patches inside a memory \cite{yi2020patch}.
Instead, CutPaste proposes memorizing a Gaussian distribution like Padim, but its performance is very similar to Padim, with the disadvantage of training an entire network \cite{li2021cutpaste}.
\\
\textbf{Normalizing Flow approaches} utilizes normalizing flows as a probability distribution estimator. During training, they learn to transform input visual features into a tractable distribution, which is used to recognize anomalies in the inference phase.
However, the disadvantage of these methods is that the memory to store the normalizing flow model and computation required for training makes it unfeasible for edge devices \cite{yu2021fastflow, gudovskiy2022cflow}. 
\\
Although many recent AD algorithms were proposed in the literature to improve performance, few works investigated how to bring these methods to real-world environments with tiny devices that have limited resources.

\begin{figure}[th]
    \centering
    \includegraphics[width=0.5\linewidth]{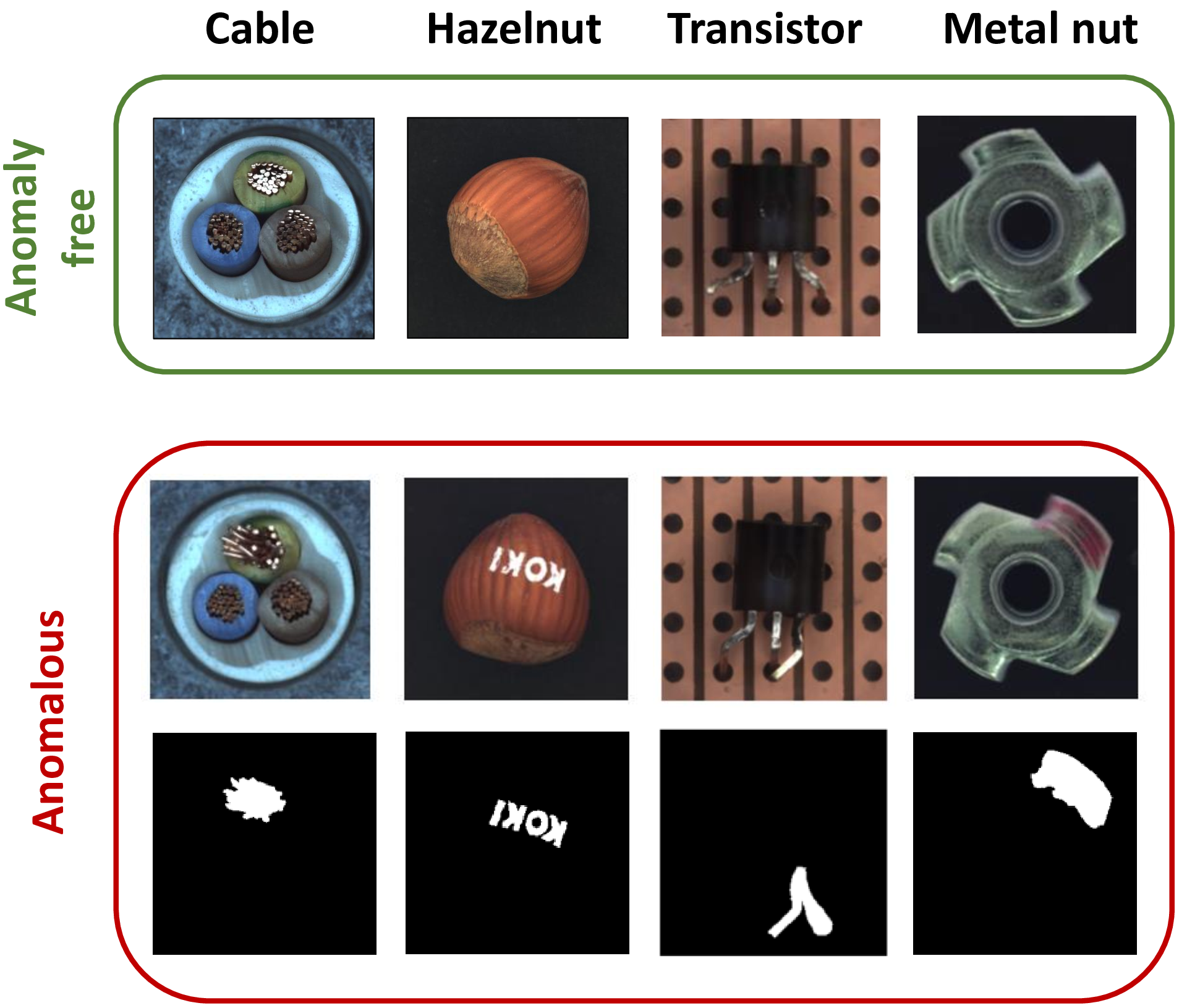}
  \caption{Image examples from the MVTec Dataset AD. Each object is shown as a normal sample (in green) and an anomalous sample (in red). }
  \label{Fig:MVTec_Dataset_AD}
\end{figure}

\subsection{Deep Learning for the Edge} \label{sec:related_work_edge}
Designing and deploying edge neural networks has been a topic that recently attracted significant attention~\cite{tinyCLAP, ancilotto2023xinet, ray2022review}. Common strategies differ on the basis of their trade-off and design principles. Neural network designs based on Neural Architecture Search (NAS)~\cite{Zoph2016NeuralAS, chen2020mnasfpn} achieve a good performance-complexity trade-off but require extensive search-space exploration. 
In NAS-based techniques, models are trained with varying hyperparameter configurations and evaluated on a held-out validation set~\cite{White2023NeuralAS}. 
This operation is very costly, as it requires sequential network training. 
Faster alternatives that concurrently train multiple networks include super-networks~\cite{Muoz2022AutomatedSG, cai2019once} and strategies based on the lottery-ticket hypothesis~\cite{Frankle2018TheLT}. 
These methods are more computationally efficient. However, they are difficult to adapt to advanced training strategies (e.g., using the network's latent representation during training). 

On the other hand, efficient neural architecture designs do not require ad hoc training and pruning strategies but rely on provably efficient operations (e.g., convolution micro-factorizations). These designs are usually parametric, and their hyper-parameters scale the computational budgets according to design-specific patterns. In this paper, we explore efficient designs based on various optimizations. 
\mobilenet/~\cite{Sandler_2018_CVPR}, MCUNet~\cite{Lin2020MCUNetTD}, PhiNet~\cite{Paissan2021PhiNetsAS} leverage the inverted residual block sequence of pointwise, depthwise, and pointwise convolutions to reduce the memory footprint of the model. Despite these designs being based on the same computational block, their scaling strategies differ, resulting in diverse performance-complexity trade-offs. \mobilenet/ scales the number of input and output channels of the convolutional block. MCUNet follows the same strategy, but removes the final network layers to reduce the minimum model footprint. PhiNet adds three scaling hyper-parameters that enable a disjoint optimization of RAM, operations, and FLASH usage by modifying the number of channels in the convolutional blocks, the depth of the network, and the expansion factor of the inverted residual block. MicroNet ~\cite{Li2021MicroNetII}, instead, reduces the number of operations of the model by proposing an efficient factorization of the depthwise and pointwise convolutions.

\section{Methodology}
\label{sec:methodology}
In this section, we present the methodology proposed in this manuscript. Specifically, Sec. \ref{subsec:general_approach} presents the general approach to bringing AD methods into real-world applications on tiny devices by changing the feature extractor from a heavy architecture to a lightweight neural network.
Then, in Sec. \ref{subsec:our_approach}, we describe our resource-efficient AD algorithm based on the STFPM approach.

\begin{figure*}[!ht]
  \centering
  
  \begin{subfigure}[b]{0.47\linewidth}
    \centering
    \fbox{
    \includegraphics[width=\linewidth]{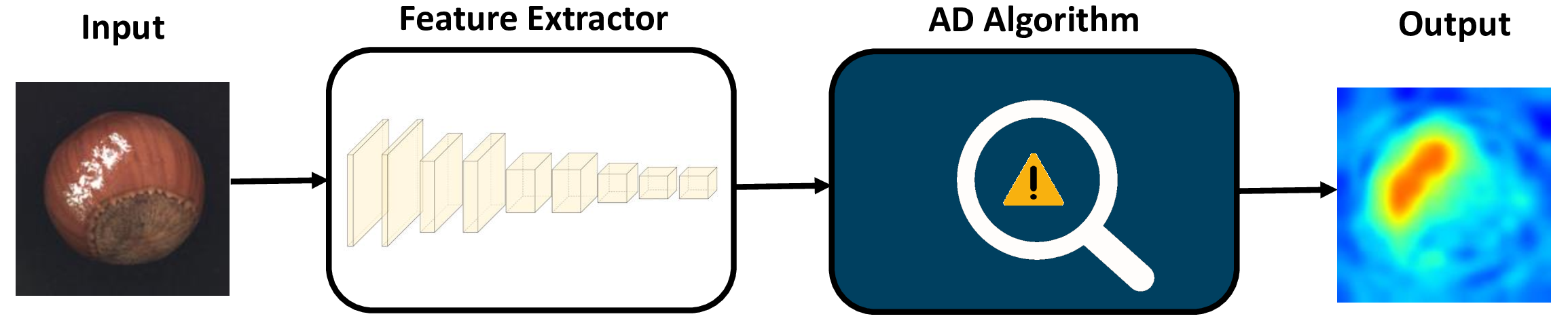}
    }
    \caption{Cloud version}
    \label{fig:scheme_Standard}
  \end{subfigure}
  \hfill
  \begin{subfigure}[b]{0.47\linewidth}
    \centering
    \fbox{
    \includegraphics[width=\linewidth]{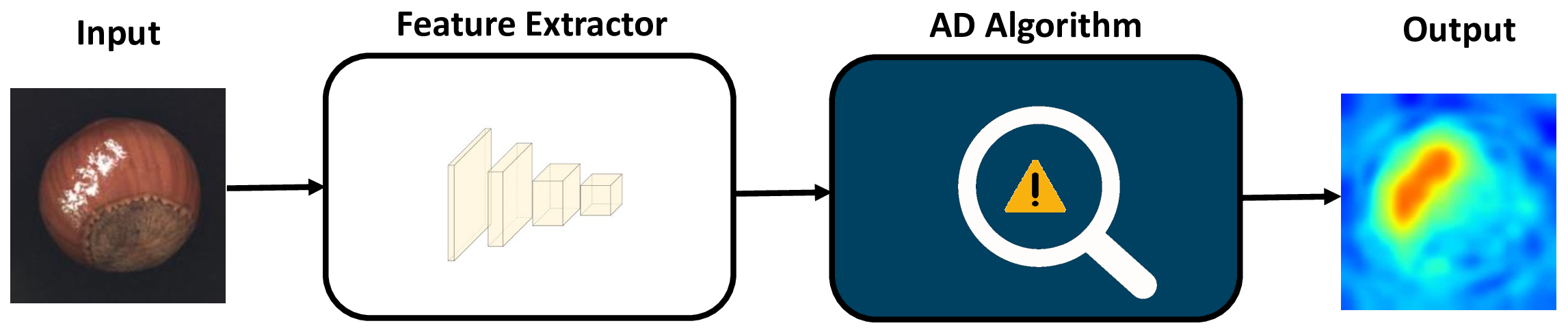}
    }
    \caption{Edge version}
    \label{fig:scheme_Edge}
  \end{subfigure}
  
  \caption{ (a) Representing the scheme for a features-based approach. Each method exploits a feature extractor, and then an AD algorithm uses such representation. The AD algorithm indicates any Features-based method such as PatchCore, Padim, CFA, STFPM, etc. (b) For the edge version, while the AD algorithm remains the same, the feature extractor is changed with a less expensive one, reducing significantly the memory and computation needed. }
  \label{fig:comparison_generic}
\end{figure*}

\subsection{General Approach}
\label{subsec:general_approach}
In the AD literature, current state-of-the-art approaches are feature-based methods, which exploit the representations produced by a pre-trained model. 
Although the proposed AD methods differ significantly from each other, a common component of all these approaches is that they are based on a feature extractor (see Fig. \ref{fig:scheme_Standard}).

AD approaches are developed and tested using large models such as \resnet/~\cite{zagoruyko2016wide}, thus without considering the challenging scenario of deploying AD for edge inference.

Therefore, to make it more feasible to deploy the AD algorithms on edge, we propose replacing the heavy feature extractor used in the AD methods with a light feature extractor (see Fig. \ref{fig:comparison_generic}).
Our first goal is to analyze the features produced by such networks and to evaluate if they can still provide enough good representations to perform the AD task.
Subsequently, following this framework, we propose a benchmark for resource-efficient VAD by evaluating several state-of-the-art AD methods by replacing computationally heavy feature extractors with light architectures and comparing several edge-oriented backbones in terms of performance and required resources.

Moreover, some AD methods could be more suitable for the edge than others.
For example, PatchCore does not have trainable weights, avoiding the costly training of a neural network. 
However, additional memory is required to store the training normal patches. 

Another advantage is that lightweight architectures typically produce smaller feature maps than larger models. This reduction in feature map size further contributes to lower memory requirements, particularly for Memory-Bank methods such as PatchCore, Padim, and CFA, where the size of the memory bank is directly related to the feature map dimensions.

Moreover, while some AD methods, like CFA and STFPM, require learning neural network weights, they involve fewer parameters and computations than larger networks. This leads to faster and more efficient training processes, which makes it easier to deploy and update VAD models on edge devices \cite{bugarin2024unveiling, xie2024iad}.

CFA uses some trainable weights, they are much less than those of the frozen feature extractor.
In contrast, STFPM does not require a memory bank; it needs to store a feature extractor plus a trainable architecture, which increases training time and inference time.
To avoid this major constraint of the STFPM and make its use on the edge feasible, we propose some modifications that significantly reduce the resource for training and inference time.

Eventually, lightweight networks allow for faster inference times. This is essential for real-time VAD applications where quick anomaly detection is critical.

\subsection{Partially Shared Teacher Student (\paste/)}
\label{subsec:our_approach}
Changing the feature extractor used in each AD method allows significant reductions in the required resources, helping them to be deployed on tiny devices. 
However, modifying each algorithm with specific considerations could help bring such methods to even smaller devices.
Specifically, in this work, we propose a modified version of STFPM, called Partially Shared Teacher-student (\paste/), which significantly reduces the required resources such as memory and computation.

\begin{figure*}[!th]
  \centering
  \begin{subfigure}[b]{0.49\linewidth}
    \centering
    \includegraphics[width=\linewidth]{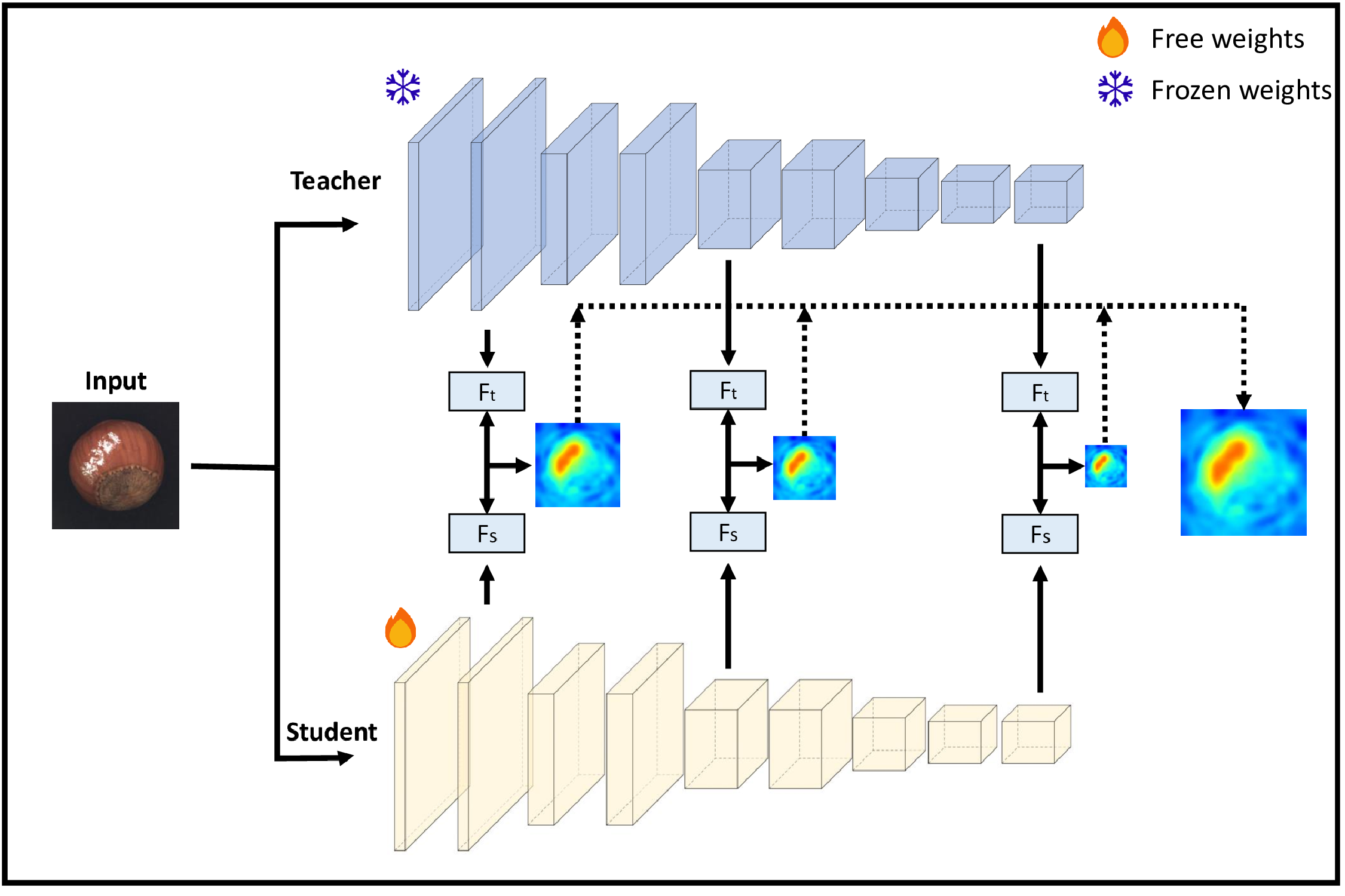}
    \caption{STFPM}
    \label{fig:STFPM_scheme}

  \end{subfigure}
  \hfill
  \begin{subfigure}[b]{0.49\linewidth}
    \centering
    \includegraphics[width=\linewidth]{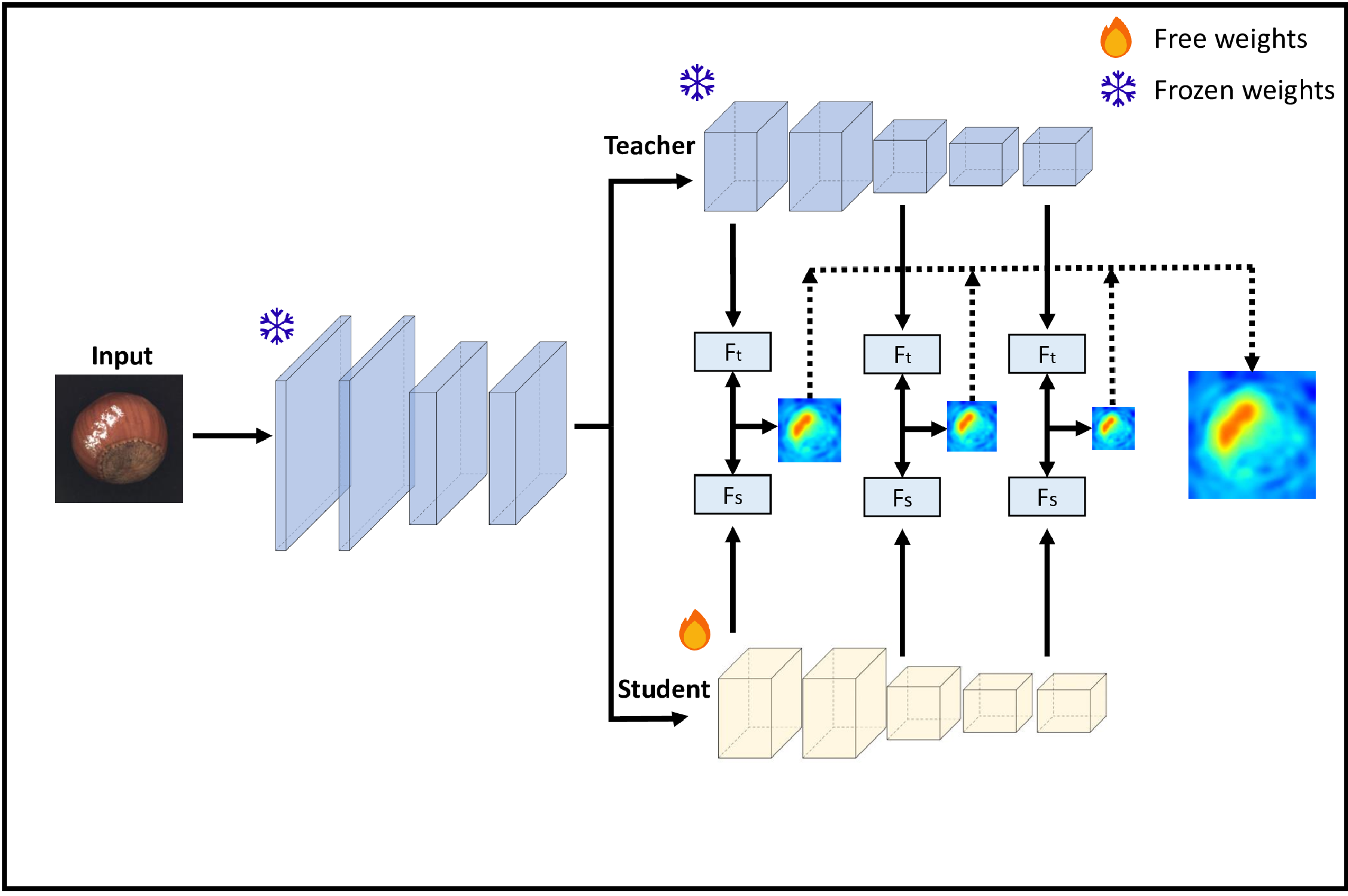}
    \caption{Paste}
    \label{fig:Latent_STFPM_scheme}
  \end{subfigure}
  \caption{Comparison between their and our approach.  It requires memorizing two architectures and performing backward on the entire architecture. It reduces the memory required for and computation resources at the minimum. }
  \label{fig:OurApproach}
\end{figure*}

The STFPM approach passes the input image through a Teacher network and a Student network. The Teacher network is typically pre-trained, while the Student network is trained to mimic the teacher's output. During the process, the features extracted at various levels from the Teacher and Student are compared (denoted as $F_T$ for the teacher and $F_S$ for the student) across different layers. 
The comparison aims to identify discrepancies between the Teacher and Student representations, which could indicate anomalies (see Fig. \ref{fig:STFPM_scheme} for an overview). 
However, one of the major drawbacks of STFPM is that it requires storing two full architectures in memory and performing backpropagation across both.
This significantly increases memory usage and computational complexity.

Our solution, \paste/, offers significant improvements over the traditional STFPM method by optimizing memory usage, reducing inference time, and lowering the computational power and RAM needed for training.
The proposed approach is depicted in Fig. \ref{fig:Latent_STFPM_scheme}.
The approach focuses on intermediate layers instead of comparing features of the first layers for both the Teacher and Student models.
The idea is based on the insight that the first layers are not fundamental for performance.
The first layers have the advantage of being the ones with more granularity, but they are also the ones with very generic features and could not be so relevant to detect anomalies, even the smallest ones.

Therefore, we assume that even if the first layers are common for teacher and student architectures, the performance using the subsequent layers should not deteriorate significantly. 
This formulation has the advantage that the first layers, which are common for teacher and student architectures, can be saved only once, saving a portion of memory since the first layers are usually the largest ones and reducing the inference time during deployment.
Moreover, since the first part of the student architecture is frozen during training, only the last part needs to be trained, significantly reducing the RAM and computational power required.
Therefore, our solution has the potential to decrease the resources needed to perform the STFPM approach on the edge, making it a more scalable and efficient solution for visual anomaly detection tasks on tiny devices.

\section{Experimental Setting}
\label{sec:experimental_setting}
This work introduces a benchmark to evaluate visual tiny anomaly detection in real-world environments, specifically focusing on resource-constrained devices (Edge computing). 
Therefore, Sec. \ref{subsec:edge_models} provides information on all edge models tested, while Sec. \ref{subsec:ad_methods} give implementation details of all the AD methods that use such backbones.
Then Sec. \ref{subsec:dataset} describes the dataset used to evaluate AD algorithms and Sec. \ref{subsec:evaluation_metrics} describes all the metrics considered to compare the AD methods.
Finally, Sec. \ref{subsec:grid_search} describes how the layers used for the feature extractor were chosen for each backbone.

To evaluate the performance of the AD methods with different feature extractor backbones, we need to make sure that the pre-trained models have all been trained on the same dataset. Specifically, we used the available weights of the models trained on ImageNet for the classification task. Moreover, every feature extractor backbone is trimmed to the last considered layer to save resources.

\subsection{Deep Learning models for Edge}
\label{subsec:edge_models}
\noindent \textbf{\mobilenet/}: available in PyTorch, we used the version pre-trained on ImageNet~\cite{deng2009imagenet} available in TorchVision.
\\
\textbf{MCUNet}: PyTorch based implementation of the MCUNet-in3 network pre-trained on ImageNet available on the official GitHub repository \cite{repo_mcunet}.
\\
\textbf{PhiNet}: PyTorch-based implementation trained on ImageNet, the source code is available in the MicroMind repository \cite{repo_micromind}. Specifically, the hyperparameters of the considered PhiNet are:  
\texttt{num\_layers=7}, \texttt{alpha = 1.2}, \texttt{beta = 0.5}, \texttt{t0 = 6}.
\\
\textbf{MicroNet}: PyTorch-based implementation of the MicroNet-m1 network pre-trained on ImageNet. The network weights are available on the official GitHub repository \cite{repo_micronet}; the architecture code has been refactored by ourselves. 
Among all the available versions of the MicroNet, PhiNet, and MCUNet models, we chose the version with the same input size and similar MACs as the \mobilenet/ model. 

\begin{table}[ht]
\centering
\caption{
Index of the layer groups used for feature extraction, for each backbone in the grid search.
\textit{Low}, \textit{Mid}, and \textit{High} refer to the depth of the layer group in the particular backbone architecture.
\textit{Equiv} refers to the layers which are equivalent in terms of \%MACs to the first three of the \resnet/ backbone.
Finally, \textit{\paste/} refers to the same layers as in the \textit{Equiv} group, but the first layers have been
shifted to account for the Partial Teacher Sharing technique. 
\textit{Last} refers to the index of the last layer of the feature extraction backbone.
}
\begin{tabular}{l|c|c|c|c}
\toprule
\multicolumn{1}{c|}{\textbf{}} & \textbf{PhiNet} & \textbf{MicroNet} & \textbf{MCUNet} & \textbf{\mobilenet/} \\ \midrule
Low   & {[4, 5, 6]} & {[1, 2, 3]} & {[3, 6, 9]}   & {[4, 7, 10]}   \\ 
Mid   & {[5, 6, 7]} & {[2, 3, 4]} & {[6, 9, 12]}  & {[7, 10, 13]}  \\ 
High  & {[6, 7, 8]} & {[3, 4, 5]} & {[9, 12, 15]} & {[10, 13, 16]} \\
Equiv & {[2, 6, 7]} & {[2, 4, 5]} & {[2, 6, 14]}  & {[3, 8, 14]}   \\
PaSTe & {[5, 6, 7]} & {[3, 4, 5]} & {[6, 10, 14]} & {[7, 10, 14]}  \\ 
\midrule
Last & 9 & 7 & 17 & 18  \\ 

\bottomrule
\end{tabular}%
\label{tab:layer_groups}
\end{table}

\subsection{AD methods}
\label{subsec:ad_methods}
\noindent \textbf{PatchCore}: tested by considering the default training and evaluation parameters \cite{patchcore_off}. The memory bank size is the default one, and the random projection applied to the feature vectors is performed with the same parameters. 
\\
\textbf{CFA}: tested by adapting the components of the method to the different feature extractor backbones. Specifically, the Patch-Descriptor network has been dynamically adapted to the dimensions of the feature vectors extracted by the considered backbones. The other training parameters, such as batch size, optimizers, and so on, are the same as the original implementation \cite{cfa_off}
\\
\textbf{Padim}: tested by considering the default training and evaluation parameters \cite{PaDiM}. The memory bank size is the default one, and the random projection applied to the feature vectors is also performed with the same parameters. 
\\
\textbf{STFPM}: all hyper-parameters are the same as the original~\cite{repo_stfpm}, except when using the MicroNet-m1 backbone, where the learning rate of SGD optimizer had to be lowered to $1/10$ the original one due to training instability, and to compensate for that the number of epochs was multiplied by $10$.
\\

\subsection{Dataset}
\label{subsec:dataset}
The MVTec Anomaly Detection (AD) dataset is a real-world dataset and is the most well-known dataset in literature to evaluate VAD algorithms \cite{MVTec}. 
This dataset encompasses ten objects and five textures, making it suitable for assessing various AD techniques' robustness and generalization capabilities. 
Specifically, the MVTec AD dataset contains over 5,000 images, encompassing 15 different object categories, such as bottles, cables, capsules, and wood (see some examples in Fig. \ref{Fig:MVTec_Dataset_AD}).

\subsection{Evaluation metrics}
\label{subsec:evaluation_metrics}

Various evaluation metrics are commonly employed to assess the performance of AD techniques.
In general, the evaluation metrics can be image-level or pixel-level.
\textbf{Image-level metrics} determine if the whole image is anomalous, while \textbf{pixel-level metrics} assess how well the model could identify the anomalous parts of the image.
For both image and pixel levels, ROC AUC and F1 metrics are usually considered.
As the main metric for assessing the anomaly segmentation performance, we decided to choose the F1 pixel level metric, which is a robust metric when there is an imbalance in pixel classes: the typical scenario in Visual Anomaly Detection where a lot of pixels are normal and only a small portion of them is anomalous. 
Furthermore, the F1 pixel level score is the strictest metric, so a high score on this metric guarantees a high score on the other metrics as well.

This work introduces a benchmark to evaluate visual tiny anomaly detection in real-world environments, specifically focusing on resource-constrained devices (Edge computing).
Therefore, we move beyond the AD performance, and report other important metrics for edge such as the AD Model memory footprint and the inference MACs (multiply-accumulate operations).
Specifically, the memory footprint represents the memory occupied by the feature extractor but also by additional components used by the AD method, such as the memory bank for PatchCore, Padim, and CFA, as well as additional architectures like the Student for STFPM and the PatchDescriptor for the CFA.

\begin{figure*}[!th]
  \centering
  \includegraphics[width=\linewidth]{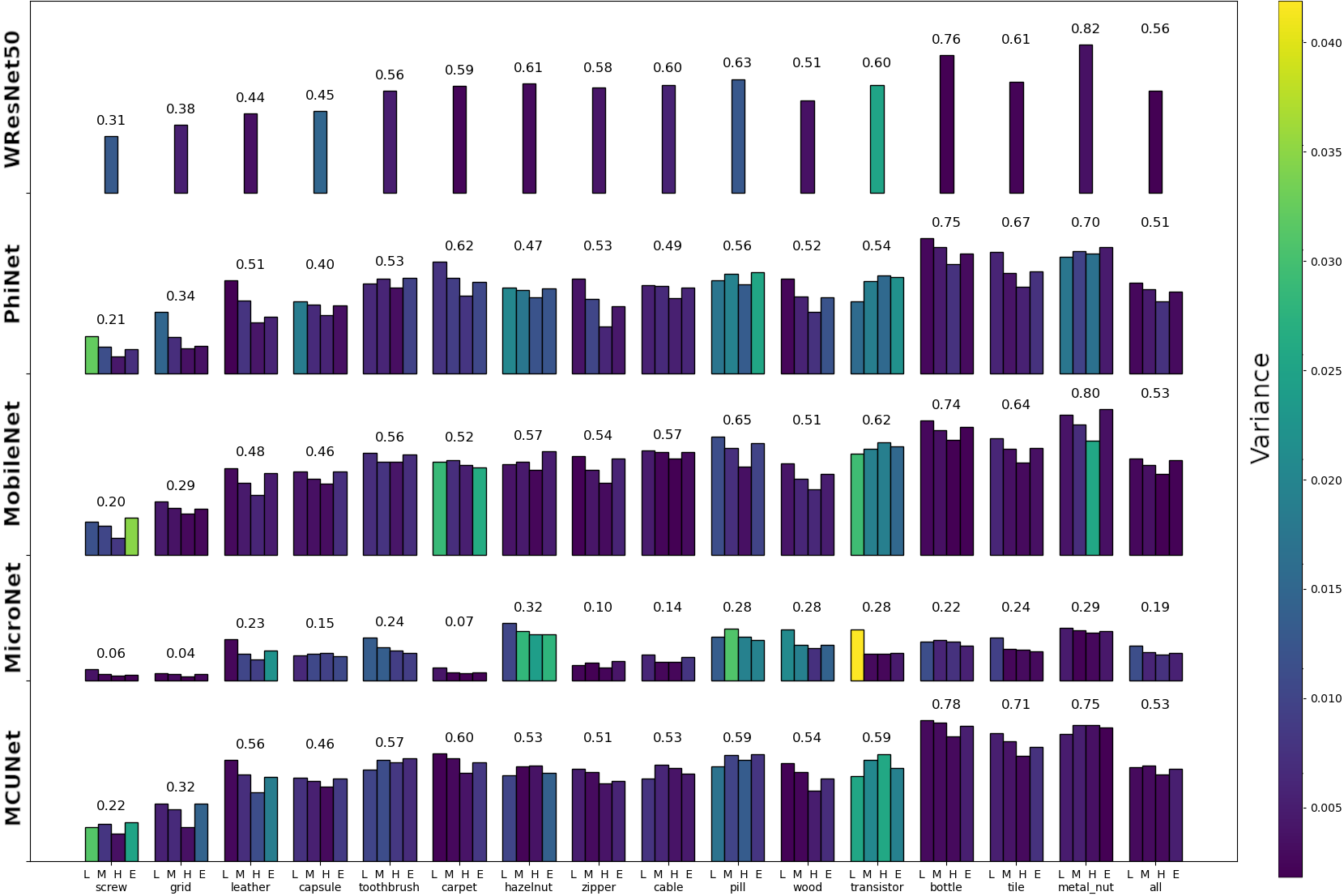}
  \caption{Overall plot of our benchmark. For every backbone, the four layers groups (L: low, M: middle, H: high, and E: equivalent,(see Tab. \ref{tab:layer_groups}) are considered for every category and are represented by a single bar. The height of the bar represents the average F1 pixel level score obtained by the different AD models using that layer group. Since every AD model is different, the color of the bar represents the variance of the F1 pixel level score. The number reported above every histogram is the F1 score of the maximum bar. The final category, named "all", represents the performance of the different layer groups on average in all the categories.}
  \label{fig:comparison_layers_category}
\end{figure*}

\subsection{Feature Extraction Layers Selection}
\label{subsec:grid_search}
Depending on the chosen layers to perform feature extraction, different performances and levels of granularity can be obtained.
For example, the first layers of every CNN extract very low-level features but with the highest granularity.
In contrast, the last layers extract high-level features that are more related to the dataset where the CNN is trained, but they also have worse granularity.
Therefore, we will evaluate the different architectures by performing a grid search on the layer groups used for feature extraction by considering both low-level and high-level layers. 

For every feature extractor, we have defined groups of layers with a \textbf{low, middle, high} depth.
The choice of the layers for every group has been defined based on the total number of layers and by considering a minimum "distance" between the layers to vary their receptive fields starting from the "center" backbone layer. For example, for \mobilenet/, which has a total of 18 layers, we defined the mid-level group of layers by considering the center layer (more or less layer with index 10), and by considering an offset of 3 layers, we considered the others: 7 and 13. The low-level and high-level layers are [4,7,10] and [10,13,16], respectively, by applying the same offset and criteria to the left and right. All the combinations of feature extractor layers considered in the experiments are reported in Tab. \ref{tab:layer_groups}. 

\resnet/, which is commonly used in AD methods, examines the features produced by three different layers, each at different depths and granularities.
Intuitively, this is a good strategy, since the anomalies vary in size. Therefore, considering features extracted from different layers simultaneously is essential to evaluate anomalies by exploiting different receptive fields. 
Therefore, in our analysis, we defined an \textit{equivalent} feature layers group, which uses layers that are equivalent
in terms of \%MACs to the heavy backbone layers considered by the official implementations of the AD models.  
For example, the first three layers of a \resnet/ use 18.39\%, 43.64\% and 80.61\% of network MACs, and the \mobilenet/ layers that use more or less the same amount of MACs are layers 3,8,4, which use 25.31\%, 43.78\% and 75.28\%.
For our method, \paste/, the chosen layers must be adapted from the original used in the equivalent layer group. Since \paste/ freezes the first layers of the network, they cannot be used, and a deeper layer group needs to be selected.
Our experiments focus on the edge-backbones, so we will not apply \paste/ to \resnet/ which is composed of just 4 layer blocks, and removing the first would lead to a steep decline in performance.
Focusing on the equivalent layers for each backbone (Tab.~\ref{tab:layer_groups}) we have to apply a shift to the selected layers in order to account for feature sharing. 
For example, the equivalent layers of \mobilenet/ are {[3, 8, 14]} and we chose to share up to the 6th layer, starting the student layer from the 7th onwards. The last selected layer inside the group must be the same since changing that would lead to an unfair comparison between the methods. The resulting feature extraction layers when using \paste/ for \mobilenet/ are {[7, 10, 14]}, where the central layer depth has also been increased to have a better spread inside the range.

\begin{table}
\centering
\caption{Summary table for comparing the original STFPM approach and our modified version for the edge called Memory-efficient STFPM.
The results here are tested using \mobilenet/ as the backbone, the layers used for feature extraction are the ones equivalent to \resnet/
\ref{tab:layer_groups} and by freezing until layer 6 for \paste/. 
While improvements in terms of memory are small, the gains in terms of inference, training computation, and training memory are significant, while obtaining very similar AD performance. }
\label{Tab:summary_table_f1_part2}
\centering
\begin{tabular}{l|S|S|S}
\toprule
                                & \textbf{STFPM} & \textbf{\paste/} & \textbf{Improvement [\%]}               \\
\midrule
\textbf{Memory [MB]  }          & \centering{5.32}      & \centering{5.11}      & \centering{3.9}           \\

\textbf{Inference [MAC] }       & \centering{454.4M}    & \centering{341.2M}    & \centering{24.9}          \\

\textbf{Training [MAC] }        & \centering{297.5M}    & \centering{198.4M}    & \centering{33.3}          \\

\textbf{RAM Training [MB] }     &  \centering{96.15}     & \centering{22.9}       & \centering{76.2}          \\

\textbf{AD Performance [F1] }   &  \centering{0.52} & \centering{0.53}      &   \centering{1.5}             \\
\bottomrule

\end{tabular}
\label{tab:comparison_stfpm_metrics}
\end{table}

\section{Results}
\label{sec:results}
Sec. \ref{subsec:benchmark_results} shows the results of our benchmark, where we use lightweight neural networks to allow AD methods to be deployed on the edge.
Then, Sec. \ref{subsec:effect_chosen_layers} investigates deeply how the chosen layers of the feature extractor affect the final performance.
Eventually, Sec. \ref{subsec:results_paste_sftpm} discusses how our novel algorithm, \paste/, reduces the resources required by the STFPM approach.

\subsection{Lightweight Neural Networks vs Standard AD Methods}
\label{subsec:benchmark_results}
As discussed in Sec. \ref{sec:related_work}, the current AD methods reached optimal results in the field.
However, there is a lack of consideration of how these methods behave in devices with constrained resources, such as edge devices.
Therefore, our first contribution is to evaluate how the main AD methods behave when considering limited resources.
To achieve this, we change the backbone used in feature-based methods, moving from a large backbone like \resnet/, commonly used in the literature, to lightweight neural networks like \mobilenet/.

In Tab.~\ref{tab:comparison_ad_approaches}, a comparison is provided in terms of performance and required resources between AD methods when using the \resnet/ as feature extractor or a \mobilenet/ with equivalent layers (a comparison table with all edge backbones is provided in the Supplementary Material \footnote{https://bitbucket.org/tinyad24/paste-sup/src/main/}).
It is fundamental to consider that each AD method has its peculiarities that cause some methods to be more memory-consuming or computing-consuming compared to others.
Therefore, the optimal AD method will be decided based not only on the AD performance but also on the resources available on the target edge device.

In general, we can see that all AD methods perform well on edge, with similar AD performance compared to the corresponding methods with \resnet/.
For example, PatchCore obtains 0.57 and 0.53 for \resnet/ and \mobilenet/ respectively. 
However, other AD methods receive even less impact by changing the underlying backbone. 
For instance, STFPM achieves the same \resnet/ performance using \mobilenet/.

Furthermore, the same or similar performance obtained with \resnet/ is achieved with a significant reduction in resources (model memory footprint and inference MACs) when using \mobilenet/.
For example, when considering memory, \mobilenet/ reduces the PatchCore memory footprint significantly from 300MB of \resnet/ to 31MB.
However, this value could still be too demanding for many tiny devices, so other methods like CFA and STFPM could be preferred with, respectively, 6.2MB and 5.3MB.
However, while CFA and STFPM appear to be the lightest among the studied approaches and with similar memory usage, when considering the inference time, the STFPM is around six times smaller, making the optimal choice for real-time applications.

In general, we provide a benchmark by evaluating several edge backbones such as \mobilenet/ MCUNet, MicroNet, and PhiNet on state-of-the-art AD methods such as PatchCore, Padim, CFA, and STFPM.
In Fig. \ref{fig:performance_tinyAD} for each Backbone and method, the results are shown, with the y-axis representing the F1 pixel level performance, the x-axis the memory (log scale), and the size of each point representing the MACs.
As is notable in the figure, all the lightweight neural networks show similar performance, though each one requires a different level of resources.
The only exception is the MicroNet architecture, which shows low results. This is because such a network, even if considered in its biggest version, is much smaller than the other edge models tested.
Therefore, while edge architectures proved to be fitted to be used to deploy AD methods on edge, a careful selection needs to be considered, since a too small network could not be able to produce enough rich representations for the AD algorithms.
Moreover, based on the resources available, some methods could be preferred to others.

In conclusion, adopting edge architectures leads to substantial memory and inference reductions. For example, STFPM achieves a 35-fold decrease in memory usage and a 4-fold reduction in inference requirements. Even more impressive is CFA, which lowers inference operations from 36.89 GMACs to 2.8 GMACs, a 13-fold reduction, and decreases memory usage by a factor of 23. Similarly, PatchCore reduces memory consumption by 9.6 times and inference demands by 4.4 times. Most notably, Padim delivers the most significant improvements, slashing memory usage from 3.72 GB to just 31.1 MB, representing a 75.4-fold reduction. 
However, even more significant is the inference reduction factor of x224. This is due to the fact that the Mahanabolois distance has cubic complexity with respect to the number of features. For WideResNet50, the features processed by Padim have a dimension of 550, while using edge architecture the feature maps are smaller, with a dimension of 62. These advancements underscore the effectiveness of edge architectures in optimizing both memory and computational efficiency. 

\begin{table*}[!ht]
\centering
\caption{We compare the considered AD methods using the same backbone, \mobilenet/, with groups of layers equivalent to WideResnet50 for feature extraction. For every AD Model, the F1 pixel level score, the memory footprint and the inference MACs are reported. With minimal lowering of F1 score, inference MACs and memory footprint of models drop dramatically.}
\resizebox{\textwidth}{!}{%
\begin{tabular}{l|SSSS|SSSSS}
\toprule
 & \multicolumn{4}{c|}{\textbf{\resnet/}}  & \multicolumn{5}{c|}{\textbf{\mobilenet/}}   \\ \midrule
 & \textbf{PatchCore} & \textbf{PaDiM} & \textbf{CFA} & \textbf{STFPM} & \textbf{PatchCore} & \textbf{PaDiM} & \textbf{CFA} & \textbf{STFPM} & \textbf{\paste/} \\ \midrule
\textbf{Total Memory {[}MB{]}} & \centering{300} & \centering{3.72G} & \centering{141}  & \centering{189.7} & \centering{31.11} & \centering{49.4} & \centering{6.16} & \centering{5.32} & \centering{5.11}   \\ 
\textbf{Inference {[}MAC{]}} & \centering{10,42G} & \centering{101,44G} & \centering{36,89G} & \centering{18.3G} & \centering{235.6M} & \centering{451M} & \centering{2.8G} & \centering{454.4M} & \centering{341.2M}  \\ 
\textbf{AD Performance {[}F1{]}} & \centering{0.57} & \centering{0.57} & \centering{0.60} & \centering{0.51} & \centering{0.53} & \centering{0.49} & \centering{0.55}& \centering{0.52} & \centering{0.53} \\
\bottomrule
\end{tabular}
}
\label{tab:comparison_ad_approaches}
\end{table*}

\subsection{How the chosen layers affect the performance}
\label{subsec:effect_chosen_layers}
After demonstrating that AD methods on edge devices can achieve performance comparable to their counterparts with heavy backbones, we investigate deeply how the layers chosen for feature extraction affect performance.
Specifically, we defined groups of layers as low, middle, and high based on their depth.
The first layers have the advantage of being the ones with more granularity, but they are also the ones with very generic features.
In contrast, the last layers extract high-level features that are more related to the dataset where the CNN is trained, but they also have worse granularity. Furthermore, we consider an \textit{equivalent} group of layers, which attempts to select layers from the lightweight backbone in the same proportion of MACs as considered in \resnet/ by the original AD methods. This group of layers actually considers a wide range of layers, covering low, middle, and high depths.
Fig. \ref{fig:comparison_layers_category} shows the results for each backbone and group of layers.
Each row represents a backbone with the groups of low, middle, high, and equivalent layers considered (highlighted with symbols L, M, H, and E), and in the column, the results for each category are shown (by averaging the AD methods), while the color represents the variance in performance with respect to the AD methods considered.
In particular, the categories shown on the x-axis are ordered from those with the smaller anomaly sizes to those with the larger ones.
For example, the screw object contains anomalies of sizes around 50 times smaller than the metal nut object.

From an initial analysis, observing the last column, $all$, gives us an idea of the performance in all categories, where similar performance is achieved for each backbone and layer, except for MicroNet, which is too limited, as discussed in the previous section.
However, a difference emerges when examining the categories individually.
The first observation is that, in general, for all the backbones, categories with larger anomalies have better performance, while the smaller ones perform worse.

In particular, while objects with bigger anomalies don't have notable differences among backbones and layers, this does not hold for smaller anomalies, like for screw and grid items.
In fact, in this case, we can observe that the \resnet/ has the best overall performance, demonstrating that when considering very small anomalies, larger backbones still have an advantage over edge backbones. 
Furthermore, for detecting small anomalies, the low-level layers play a crucial role because they have a very small field of view and because of the low-level features (such as angles, lines, and so on) that are extracted from them. 
Among the low, middle, and high groups, the low group of layers is better than the other groups for every tiny backbone, as shown in Fig. \ref{fig:comparison_layers_category}. 

However, the best results for small anomalies are usually observed when the analysis is considered the equivalent layer group as in the case of \mobilenet/ and MCUNet backbones.
This suggests that the additional information of middle-level and high-level layers can improve the detection, though considering low layers remains fundamental.
Therefore, in general, a good solution to identify both small and large anomalies seems to involve selecting a group of feature extraction layers that contains a combination of low-, middle- and high-depth layers, as originally considered for \resnet /. 

\subsection{\paste/}
\label{subsec:results_paste_sftpm}

Although the solution to use lightweight networks investigated in this work makes it possible to significantly reduce the required resources for many VAD methods, it is clear how further work should be taken into account to further optimize the resources required by each method.

For this reason, in this work, we propose an efficient version of STFPM called \paste/.
\paste/ works by reusing the same weights of the first layers of the teacher for the student architecture to bypass computation on these since we can \textit{copy} the features extracted by the teacher and \textit{paste} it as the input of the student network.
The first layers are usually the ones with the fewest weights and the most expensive in terms of backward-pass computing and RAM during training.
As is visible from Tab.~\ref{tab:comparison_stfpm_metrics}, where we compare STFPM with \paste/ using \mobilenet/ as the backbone,
the inference time is reduced by 24.9\% compared to STFPM, from the original 454.4MMAC to the 341.2MMAC of \paste/.
Instead, the memory footprint of the model does not see great improvement and is reduced by 3.9\%, from the original 5.32MB to 5.11MB.

Moreover, during training, the computation is reduced from the original 297.5MMAC of STFPM to 198.4MMAC of \paste/, with a reduction of more than 33\%.
In addition, when considering the RAM usage required by training, our method requires only 22.9MB compared to 96.15MB of STFPM, with a reduction of 76.2\%.
Therefore, our method can significantly reduce the required resources, making it more viable for deployment at the edge compared to STFPM.

\begin{figure}[!ht]
  \centering
  \includegraphics[width=0.6\textwidth]{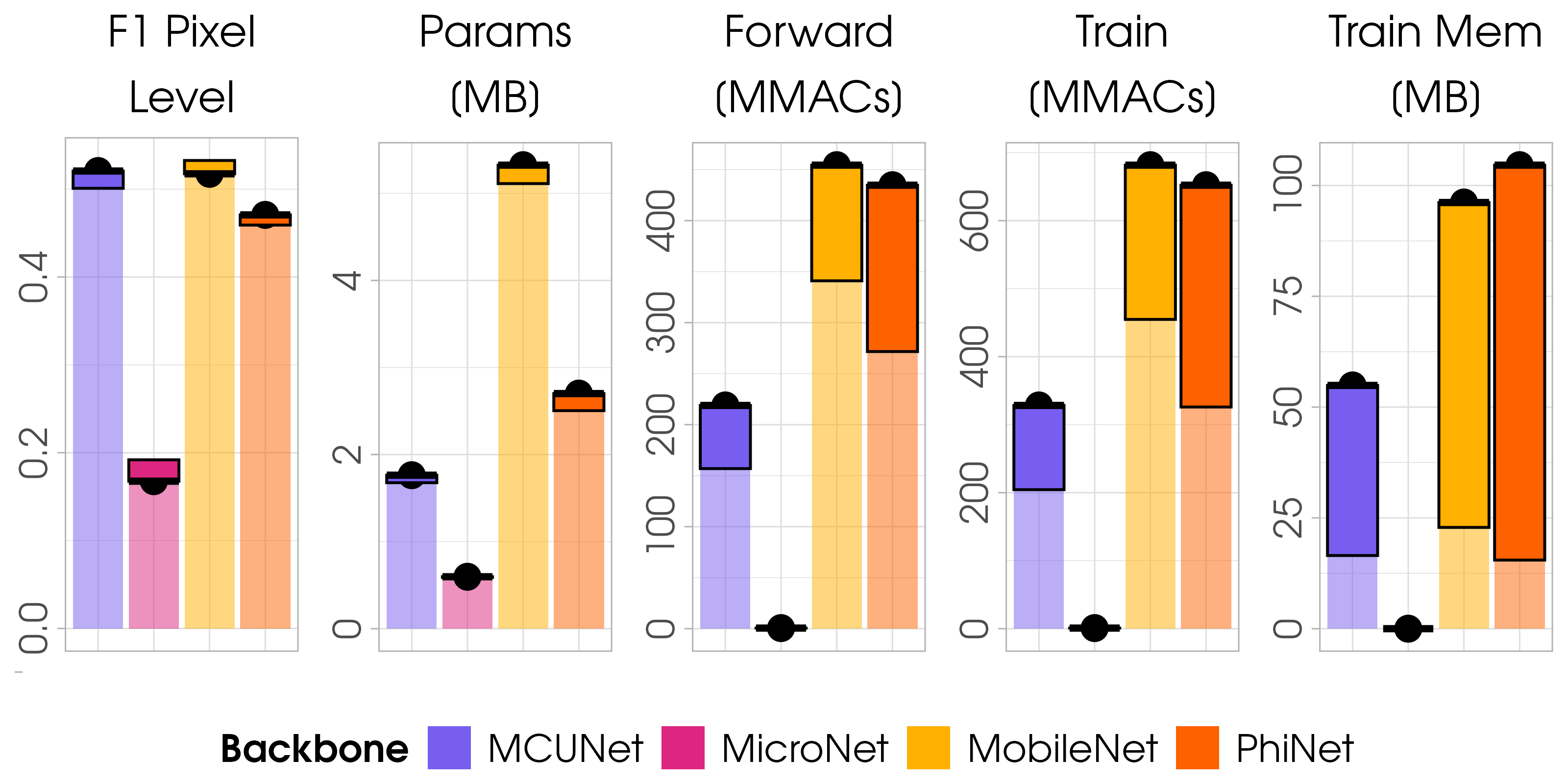}
  \caption{Difference between \paste/ and original STFPM method. For each backbone, we compare them as performance (F1), Params (MB), Inference (MMAC), Training (Million MACs), and RAM (MB).
  Micronet obtains low values making it hardly visible in the plot.
  }
  \label{fig:paste_vs_stfpm}
\end{figure}

Similar improvements are achieved for the \mobilenet/, PhiNet, and MCUNet backbones.
Fig.~\ref{fig:paste_vs_stfpm} shows a clear improvement in terms of computing and RAM memory resources, while having almost no negative impact on performance, which actually improves in some backbones.

Instead, MicroNet shows limited gains. The first reason is due to the results on the MicroNet backbone, which are not comparable to the other backbones, where its extremely compact size already sits ad its resource-efficiency limits. This is because its original layer configuration of [2, 3, 5] translates to [3, 4, 5] when adapted to \paste/, resulting in the freezing of few layers. Consequently, the potential for improvement is minimal with such limited layer adjustments.

These advantages are achieved while achieving similar performance or, in the case of some backbones like \mobilenet/, even better performance. Indeed, with \mobilenet/, STFPM achieves 0.52 f1 pixel-level compared to 0.53 of \paste/.
Furthermore, the performance for small anomalies appears to be slightly improved with \paste/ (see the results of the \paste/ approach for each category and the backbone in the Supplementary Material \footnote{https://bitbucket.org/tinyad24/paste-sup/src/main/}).
This is justified by the fact that the first layers are important, but choosing layers too close to the input could be damaging, further motivating the freezing of such layers in \paste/.

\section{Conclusion}
\label{sec:conclusion}
Visual Anomaly Detection is an important field that aims to identify anomalous images and the specific parts inside the image that are abnormal.
This is performed with an unsupervised paradigm, avoiding the costly label collection phase, especially for the pixel-level granularity.
This work provides a benchmark for TinyAD, where visual anomaly detection is studied for resource-constrained devices. This work impacts many domains, such as manufacturing, medicine, and autonomous vehicles.
However, edge devices often have limited resources, making it challenging to deploy complex deep learning models typically used in VAD.
The benchmark uses lightweight neural networks, such as \mobilenet/ and PhiNet, and explores various anomaly detection approaches, including PatchCore, CFA, Padim, and STFPM, analyzing their suitability for edge devices.
Specifically, several edge architectures are implemented, proving their ability to significantly reduce memory and computation constraints, making AD algorithms more feasible for deployment on the edge in small devices.

In addition, we introduce a novel algorithm called \paste/, which addresses the memory and computational limitations of the traditional STFPM method, making it more scalable and efficient for edge deployment.
The idea is that initial layers are not so fundamental in the final performance. Therefore, these weights are frozen and used for both the teacher and the student architecture.
\paste/ can reduce more than half the training time and more than four times the RAM for training while obtaining the same performance and reducing the inference time by 30\%.
Evaluating all the methods and architectures on the well-known MVTec AD Dataset proves the feasibility of AD algorithms for edge and the superiority of memory-efficient STFPM to STFPM.

While this work proved the feasibility of AD approaches on the edge, there are several promising research directions.
For example, while this work significantly reduced the memory and processing power usually considered for Visual Anomaly Detection, further work is necessary to reduce the resources consumed to run AD algorithms on tiny devices even further.
Moreover, a further insight from this work is that these methods, in the context of tiny AD, struggle when considering objects with very small anomalies, such as screws. Therefore, an interesting research direction will be to improve the performance of these objects.
Furthermore, we don't know how edge architectures behave in modified scenarios compared to larger architectures. For instance, we do not know how they can behave in noisy settings or when in the presence of a data stream.

\bibliographystyle{IEEEtran}
\bibliography{references}

\end{document}